\documentclass[times,referee,twocolumn,final,authoryear]{elsarticle}

\usepackage{framed,multirow}

\usepackage{hyperref}

\usepackage[table]{xcolor}

\usepackage{amssymb}
\usepackage{latexsym}

\usepackage{url}
\usepackage{hyperref}

\usepackage{tikz}
\newcommand{\dashedarrowtikz}{\raisebox{2pt}{\tikz{\draw[<-,black,dashed,line width = 0.5pt](0,0) -- (6mm,0);}}}
\newcommand{\normalarrowtikz}{\raisebox{2pt}{\tikz{\draw[->,black,solid,line width = 0.5pt](0,0) -- (6mm,0);}}}

\definecolor{lightgray}{gray}{0.9}
\definecolor{newcolor}{rgb}{.8,.349,.1}

\usepackage[utf8]{inputenc}

\usepackage[english]{babel}
\usepackage{amsmath,amssymb,amsfonts}
\usepackage{graphicx}
\usepackage{textcomp}
\usepackage{xcolor}
\usepackage[tight,footnotesize]{subfigure}
\usepackage{hyperref}
\usepackage{float}
\usepackage{multirow}
\usepackage[ruled,vlined]{algorithm2e}
\usepackage{adjustbox}

\def\Figref#1{Figure~\ref{#1}}

\def\eqref#1{equation~\ref{#1}}
\def\1{\bm{1}}

\DeclareMathAlphabet{\mathsfit}{\encodingdefault}{\sfdefault}{m}{sl}
\SetMathAlphabet{\mathsfit}{bold}{\encodingdefault}{\sfdefault}{bx}{n}

\usepackage{siunitx} % Required for alignment
\usepackage{array,booktabs}
\usepackage{url}
\usepackage{xcolor}
\definecolor{newcolor}{rgb}{.8,.349,.1}

\journal{arxiv.org}

\begin{document}

\begin{frontmatter}

\title{Attention-based fusion of semantic boundary and non-boundary information \\ to improve semantic segmentation}

% Group authors per affiliation:
\author[addressufba,addressufma]{Jefferson Fontinele}
\ead{jefferson.font@ufba.br}
\author[addressufba]{Gabriel Lefundes}
\ead{gabriel.lefundes@ufba.br}
\author[addressufba]{Luciano Oliveira\corref{cor1}}
\ead{lrebouca@ufba.br}
\cortext[cor1]{Corresponding author: Luciano Oliveira.}
\address[addressufba]{Intelligent Vision Research Lab, \\ Universidade Federal da Bahia, Bahia, Brazil}
\address[addressufma]{Universidade Federal do Maranhão, Maranhão, Brazil}

\begin{abstract}
This paper introduces a method for image semantic segmentation grounded on a novel fusion scheme, which takes place inside a deep convolutional neural network. The main goal of our proposal is to explore object boundary information to improve the overall segmentation performance. Unlike previous works that combine boundary and segmentation features, or those that use boundary information to regularize semantic segmentation, we instead propose a novel approach that embodies boundary information onto segmentation. For that, our semantic segmentation method uses two streams, which are combined through an attention gate, forming an end-to-end Y-model. To the best of our knowledge, ours is the first work to show that boundary detection can improve semantic segmentation when fused through a semantic fusion gate (attention model). We performed an extensive evaluation of our method over public data sets. We found competitive results on all data sets after comparing our proposed model with other twelve state-of-the-art segmenters, considering the same training conditions. Our proposed model achieved the best mIoU on the CityScapes, CamVid, and Pascal Context data sets, and the second best on Mapillary Vistas.
\end{abstract}


\end{frontmatter}

% \linenumbers

\section{Introduction}

%Semantically segmenting an image consists of assigning a meaningful label to each one of its pixels. 

Many vision-related applications rely on semantic segmentation to evaluate visual contents and perform decision making. For example, autonomous driving \citep{Cordts2016Cityscapes, MVD2017}, agricultural automation \citep{santos2020grape}, medical image analysis \citep{bakas2017advancing}, remote sensing imagery \citep{maggiori2017dataset}, and the human parsing task \citep{deephumanparsing} usually leverage the semantics of pixel-level labels to obtain valuable information such as an object class, location, or shape. 

As with many vision-based tasks, the current state-of-the-art in semantic segmentation heavily explores convolutional neural networks (CNN), being DeepLabV3 \citep{deeplabv3}, DeepLabV3+ \citep{deeplabv3plus}, and PSPNet \citep{pspnet} some of the best architectures in the current literature. In particular, fully convolutional networks (FCN) \citep{fcn}, which are usually used as the main branch of the CNN-based segmenters, have achieved high-quality results as demonstrated by several works \citep{deeplabv3plus, pspnet, rota2018place, ronneberger2015unet, badrinarayanan2017segnet}. FCN-based models use pre-trained classification networks to obtain a score-map that relates classes to each pixel of the input image. During this process, the classification networks downsample the map of features, progressively reducing the spatial resolution of the original input. The information loss that ensues from this downsampling is particularly relevant to semantic segmentation. The decrease of resolution causes the loss of detail, resulting in lower accuracy for smaller objects and boundary regions in general.

\begin{figure}[t]
    \centering
    \subfigure[]{\includegraphics[width=.24\textwidth]{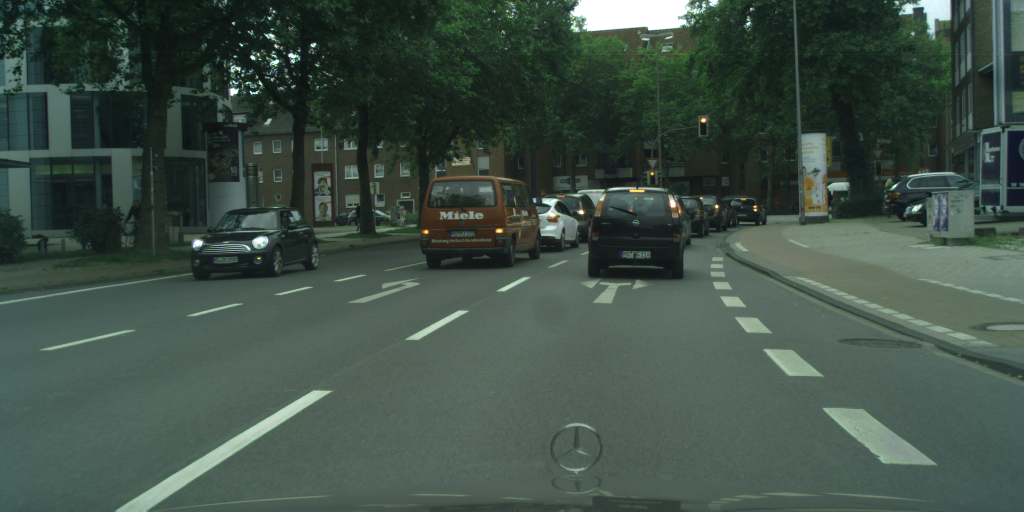}}
    \subfigure[]{\includegraphics[width=.24\textwidth]{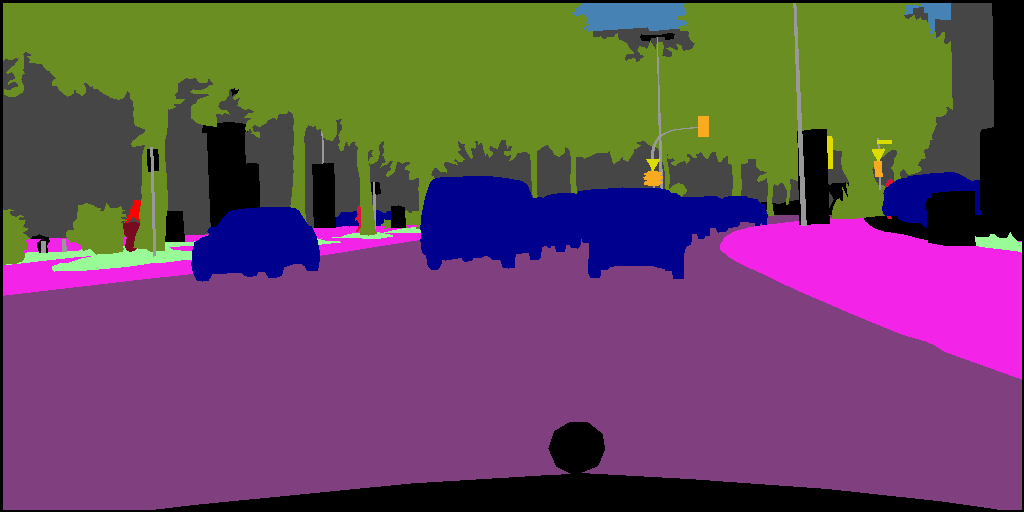}} \\
    \subfigure[]{\includegraphics[width=.24\textwidth]{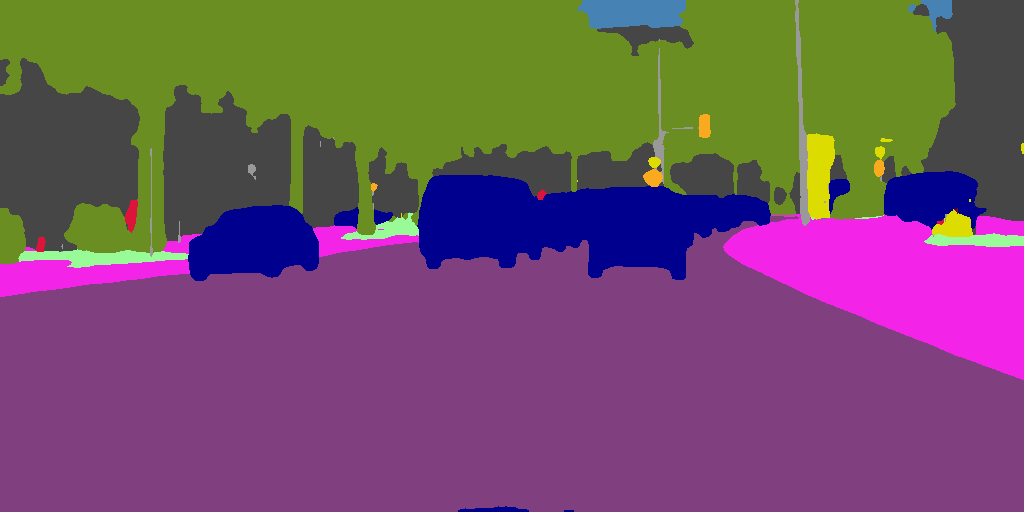}}
    \subfigure[]{\includegraphics[width=.24\textwidth]{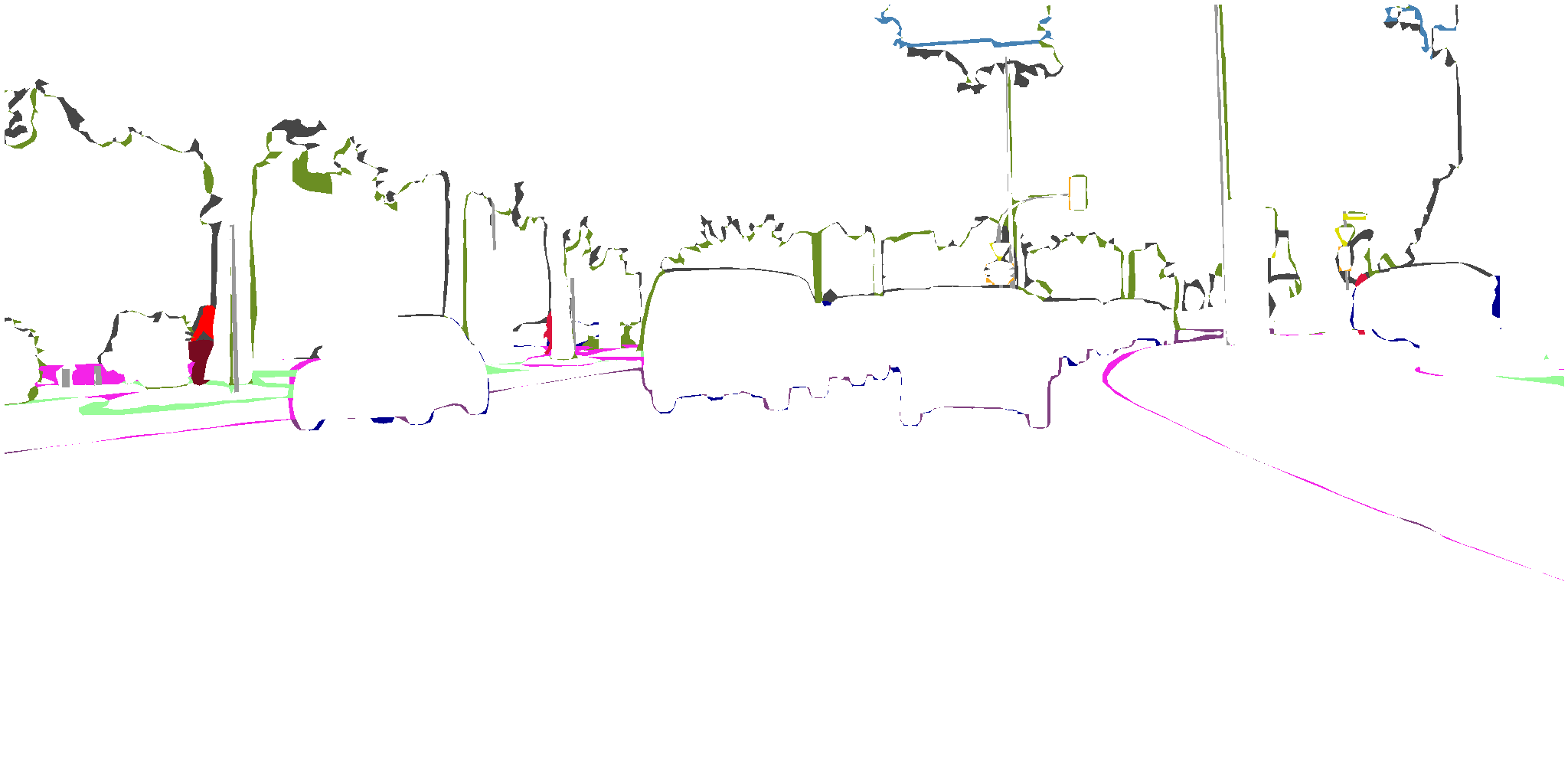}}
    
    \caption{How important are the boundaries in the process of image segmentation. (a) An image sample from Cityscapes \citep{Cordts2016Cityscapes}, (b) the ground truth, (c) the masks predicted by DeepLabV3\citep{deeplabv3}, and (d) difference between (b) and (c).}
    \label{fig:eval_quantitative}
\end{figure}

%Some works have tried to address the higher classification error of pixels located near the boundaries. For example, Takikawa \textit{et al.} \citep{gscnn} embedded a boundary detection algorithm into an FCN-based model, where the former was responsible for identifying image locations where pixel intensities changed abruptly. Similarly, Zhen \textit{et al.} \citep{joint-semantic} appended a semantic boundary detection (SBD) stream into their model in order to identify the contours of objects belonging to semantic classes. Takikawa \textit{et al.} \citep{gscnn} and Zhen \textit{et al.} \citep{joint-semantic} based their approaches on the possibility that the proper detection of boundary could improve the results of semantic segmentation. Another important point to consider is that 

Commonly segmentation models face problems in segmenting the boundaries and their surroundings. \Figref{fig:eval_quantitative} shows the result of the semantic segmentation obtained by a DeepLabV3~\citep{deeplabv3} network compared to the ground truth. The difference between ground truth (\Figref{fig:eval_quantitative}(b)) and prediction (\Figref{fig:eval_quantitative}(c)), depicted in \Figref{fig:eval_quantitative}(d), indicates that a large part of the error attributed to the model is related to boundary regions, further reinforcing the hypothesis that semantic boundaries should be included as part of the problem modeling. To minimize that problem, one could use a semantic boundary detection (SBD) module \citep{casenet, devil-edges, dff, ma2020multi}. By adding an SBD module to an FCN segmentation architecture, a new relevant challenge is introduced: \textit{How to determine adequate pixel context from boundary and segmentation information separately?} A promising avenue of investigation to address this question is to use attention gates, which are usually explored to combine low-level features and high-level features in deep architectures \citep{dualattention2}. In our work, we propose a novel approach that uses attention gate to help separate boundary and non-boundary information, seeking for the improvement of the semantic segmentation results. We investigate whether the use of semantic segmentation and SBD, combined through an attention model, can contribute to better results in semantic segmentation. The rationale is that it is possible to separately learn from different contexts by using attention models, also jointly training the model in an end-to-end network. We argue that the pixels belonging to the object boundaries have different characteristics when compared to the other image pixels, and this difference is mainly related to the context of border pixels containing features from both objects separated by that border.

%Our approach consists of using three main modules as depicted in Fig. \ref{fig:model_complete}. The first module is a semantic segmentation network, and the second one is an SDB network. Both of these share the same backbone, with the SBD using different stages of it to perform it's particular task. These two networks are connected through the third and last module, which is based on attention and similar to the one found in \citep{gscnn}, combining the semantic segmentation with the SBD module's output.

\begin{figure}[t]
     \centering
     \includegraphics[width=0.5\textwidth]{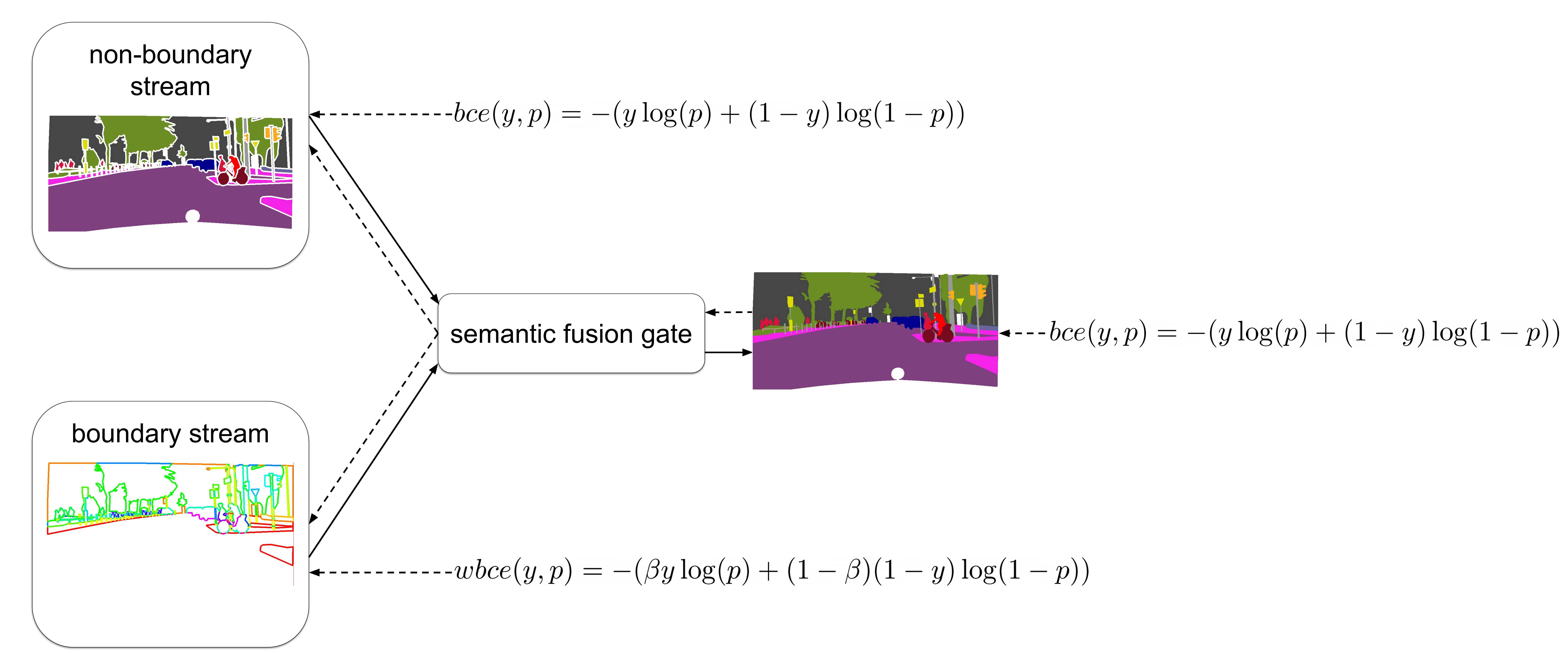}
     \caption{Our proposed Y-model is comprised of three parts: Non-boundary stream, boundary stream, and a semantic fusion gate. The semantic fusion gate is an attention gate in charge of combining the network signal  of the boundary and non-boundary streams in the forward step (segmentation), while splitting the error signal in the backward step (training).}
     \label{fig:y_model}
 \end{figure}

%($\normalarrowtikz$) ($\dashedarrowtikz$)

\Figref{fig:y_model} illustrates our approach to the problem of semantic segmentation. We call it Y-model mainly because of the shape of the network. The semantic fusion gate is an attention gate in charge of combining the network signals ($\normalarrowtikz$) of the boundary and non-boundary streams in the forward step (segmentation) while splitting the error signals ($\dashedarrowtikz$) in the backward step (training). Each stream and and the top of the attention gate have its own loss function. All losses are used to optimize the entire network all at once in an end-to-end way. We performed experiments to validate our proposal using four publicly available data sets, that are: Cityscapes~\citep{Cordts2016Cityscapes}, Mapillary Vistas~\citep{MVD2017}, CamVid~\citep{BrostowSFC:ECCV08_camvid}, and Pascal Context~\citep{mottaghi_cvpr14_pascal_context}. Cityscapes, Mapillary Vistas, and CamVid are composed of traffic images from the driver's viewpoint, while Pascal Context is a data set that presents a diversity of scenes, without theme restriction on the class sets.

The data sets used for experimentation are all burdened by the class imbalance problem. For instance, in Cityscapes, Mapillary Vistas, and CamVid, the volume of pixels belonging to the class ``road'' is much higher in scale than the other classes in any given scene (due to the inherent nature of traffic-related footage). To reduce the impact that unbalanced data may have, works such as in \citep{gscnn} and \citep{video-pro-relaxation} proposed methods that attempt to enforce uniform class distribution by selecting particular crops from the original image during training. In \citep{gscnn}, classes, which have a less overall frequency in the data set, are selected, assuring that whenever these classes are present in the image, they are also present in the selected crop. In \citep{video-pro-relaxation}, centroids are computed from the image segments and then the crop is selected by considering a uniform sampling of these centroids. In our work, we propose a new method for crop selection in unbalanced data sets. Unlike \citep{gscnn} and \citep{video-pro-relaxation}, our method assigns a weight value to each pixel from each image on the data set. These weight values are inversely proportional to the class frequencies in the overall data set. This approach makes it possible to perform crops that minimize the class unbalance issue on the training data by considering the value of each pixel of the image during crop selection.

\subsection{Contributions}

The main contributions of this work are summarized as follows: A new model comprised of non-boundary and boundary streams that divides the tasks of semantic segmentation and semantic boundary detection by means of a Y-model and an attention gate, and a new strategy for uniform crop selection that efficiently reduces the imbalance of the classes in the training data sets

\section{Related work}

Here we briefly go over the relevant published works on topics that directly or indirectly impacted our proposed model. They are semantic segmentation, boundary detection, semantic boundary detection, and attention models. 

\paragraph{Semantic segmentation} FCN-based models usually own three modules: A feature extractor that uses an image classification network (backbone), without the dense layers at the end; a module that tries to expand the receptive field through dilated convolutions \citep{yu2015multi, deeplabv3, yang2018denseaspp}, large convolutions \citep{peng2017large}, and image pooling \citep{pspnet, liu2015parsenet}; and a decoder in charge of combining the extracted features while performing pixel-wise classification to produce the segmentation map. One significant issue of FCN-based approaches is related to the spatial representation of the extracted features. The downsampling stages of the classification network reduce the spatial resolution of the features, effectively attenuating the object boundaries in the input image. Due to this effect, FCN-based segmentation networks usually produce results with less detailed boundaries when compared to the ground truth.

Some works manage to improve accuracy on the boundaries by using features from different stages of the backbone \citep{fcn, unet++, deeplabv3plus}. The rationale is that features from the early stages have better spatial representation since they are not as downsampled as features from later stages. Models such as FCN-8~\citep{fcn} and DeeplabV3+~\citep{deeplabv3plus} are examples of the networks that improve their overall results by exploiting features of the initial stages of a ResNet network~\citep{he2016deep-resnet}. However, even when combining several stages of the backbone, a significant part of the classification errors in segmentation models still occur due to failures in boundary classification, as illustrated by \Figref{fig:eval_quantitative}, where the difference between the segmentation result and the ground truth is mostly defined by incorrectly-classified boundaries. These results indicate that networks trained solely for semantic segmentation have difficulty in segmenting the borders of the objects. Simply using various stages of the backbone is not enough to completely overcome the problem of poor classification rate on the boundaries.

\begin{figure*}[t]
    \centering
    \includegraphics[width=0.9\textwidth]{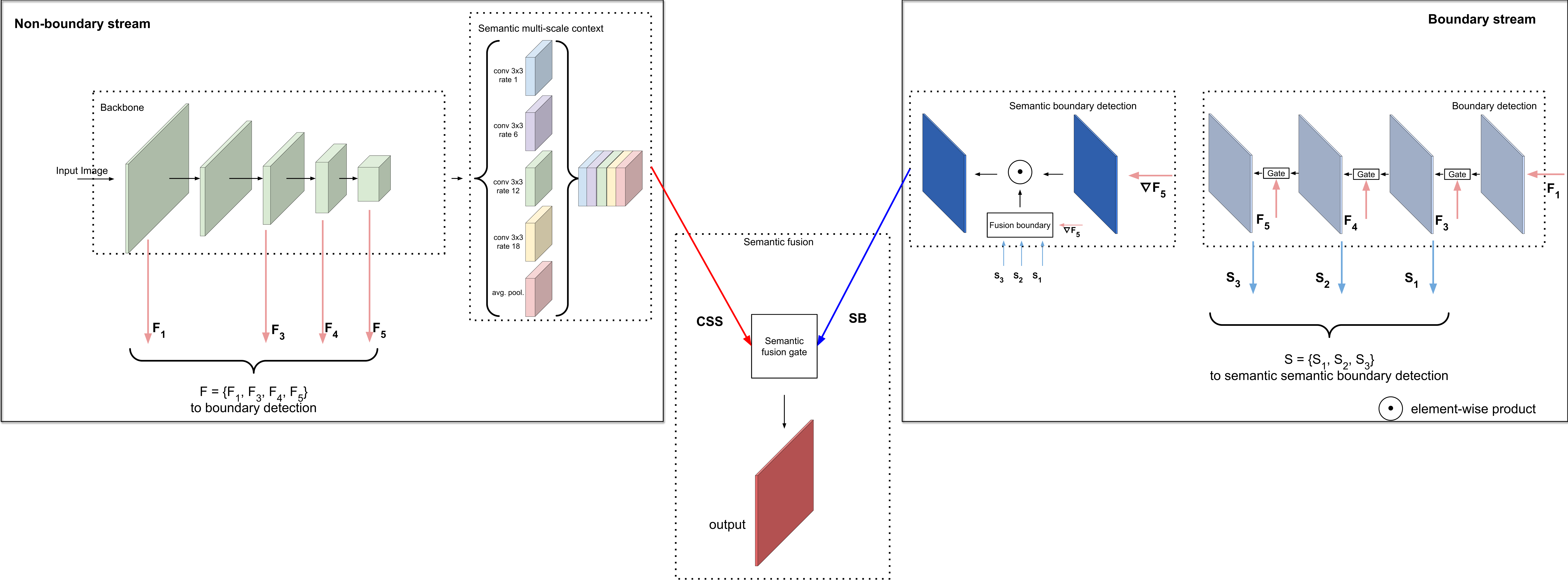}
    \caption{Detailed architecture of our Y-model. The input image is fed to a backbone that generates a set of features maps (\textbf{F}) with different scales. The feature maps \textbf{F} are used in the boundary stream to obtain a set of features \textbf{S}. Each element in \textbf{S} represents the boundary from the features of different scales in \textbf{F}. The semantic boundary detection module combines \textbf{$F_5$} and \textbf{S} to obtain the semantic boundaries (\textbf{SE}), which combined with the coarse semantic segmentation (\textbf{CSS}) through the attention gate output the segmented image.}
    \label{fig:model_complete}
\end{figure*}

\paragraph{Boundary detection} 
Some methods explore deep networks to detecting boundary \citep{xie2015holistically, redn} and semantic boundary \citep{simultaneous}. The results obtained by these methods surpass those based on hand-crafted features. For instance, Takikawa et. al. \citep{gscnn} use boundary detection to improve semantic segmentation by creating a branch that combines features extracted from several stages of a ResNet~\citep{he2016deep-resnet} network. The feature learning is then split through a gate-based model into two different streams, one for the boundaries and the other for semantic segmentation. 

%Não citaria esses FDPs... kkkk
%The results reported by the GSCNN~\citep{gscnn} network show that jointly learning a boundary detector seems to improve semantic segmentation, as it approximates the segmentation derived from the actual boundaries to the other elements of the scene. 

%%% 
%In \citep{ding2019boundary}, Ding et. al. use two streams to separate the tasks of boundary detection and semantic segmentation. The two streams share the same information stream, allowing each specializes in performing a specific task. In our experiments, we propose a two-stream approach similar to the one introduced by GSCNN \citep{gscnn}, but unlike \citep{gscnn} we considered a semantic boundary detection and a non-boundary detection in a new added stream.

\paragraph{Semantic boundary detection} More recently, methods that detect boundaries for each semantic class \citep{casenet, devil-edges, dff, ma2020multi} have obtained better results than those that perform general boundary detection. This motivated us to use semantic borders as part of our method. One should note that those works do not perform semantic segmentation, but only boundary detection.

\paragraph{Attention models} To build a multi-task neural network, there is usually the need to compound multiple loss functions and a clear strategy as to how to combine the multiple tasks. In \citep{casenet}, the use of a specific loss function based on binary cross-entropy is proposed as a way to deal with boundary classification. In our experiments, however, we have noticed that when combining regular semantic segmentation features with boundary features, the training gradient ends up attenuating the signal coming from the boundary detection stream. 

%In GSCNN~\citep{gscnn}, a Canny~\citep{canny1986computational} filter is used as an input to the network as a way to reinforce the signal. Additionally, a particular loss function is used to enhance the boundaries of the segmentation network along with the boundary detector. The boundary stream is joined with the segmentation stream to make up the final segmentation map. In our work, we explored the use of an attention model for combining and selecting boundary features, so that the undesired attenuation in the losses of boundary detection and segmentation is minimized.

\section{Fusion of boundary and non-boundary information through an attention gate}

We explored an attention-gate approach, explicitly modeling object boundaries as a way to improve semantic segmentation. To do so, we conceived a Y-model (refer to \Figref{fig:y_model}) defined by three components: A \textbf{non-boundary stream}, which produces a coarse segmentation for the input image; a \textbf{boundary stream}, responsible for processing the input features to perform border detection; and a semantic fusion gate (attention model), which integrates both streams to obtain the final semantic segmentation output. \Figref{fig:model_complete} shows in detail our complete proposed architecture. The streams are each one split into two different processing steps. The \textbf{non-boundary stream} is represented by the \textit{backbone} and \textit{semantic multi-scale context} modules, while the \textbf{boundary stream} is represented by the \textit{boundary detection} and \textit{semantic boundary detection} modules. Both streams are processed in parallel, and their outputs are combined to produce the final output via a \textbf{semantic fusion gate}.

\subsection{Non-boundary stream}

The non-boundary stream is responsible for performing a preliminary semantic segmentation step on the input image. As previously mentioned, we based our Y-model in an FCN-based network. These types of networks consist of three modules: The first is a feature extractor, which uses an image classification network as backbone. The one-by-one convolution, usually placed at the end of the network for class prediction, is excluded; the second module improves the context representation of each pixel by using dilated convolutions \citep{yu2015multi, deeplabv3, yang2018denseaspp}, large convolutions \citep{peng2017large}, and image pooling \citep{pspnet, liu2015parsenet}; the last and third module is the decoder. The decoder is responsible for combining the extracted features and performing pixel classification to produce the segmentation map. The backbone used by our proposed model relies in five stages as illustrated on the top-left part of \Figref{fig:model_complete}. We applied an FCN-based network that takes as input an image $I \in \mathbb{R}^{H \times W \times C }$ with height $H$ and width $W$, and the number of channels $C$ that outputs a map $M$ of dimensions $H \times W$. A coarse semantic segmentation ($\mathbf{CSS}$) map contains the class prediction scores for each pixel in the image, which is ultimately refined by the output of the boundary stream.

\paragraph{\textbf{Backbone}} As the base feature extractor for the non-boundary and boundary streams, we used a Dilated ResNet~\citep{dilated-resnet} (with 50 and 101 layers) and a WideResNet~\citep{wideresnet} (38 layers), in our experiments. For the Dilated ResNet, we used the default dilation rates of 2 and 4 in blocks 4 and 5 of our network, respectively, producing an output with stride 8. This design choice follows the suggestions in~\citep{dilated-resnet}. For the WideResNet, we used a dilation rate of 2 for block 3, and a dilation rate of 4 for the subsequent blocks. This final output is achieved with stride 4. The feature extractors structured with dilated convolution increment the feature output shapes, improving the image representation for the segmentation task. 

\paragraph{\textbf{Semantic multi-scale context}} We used a multi-scale representation for the context. This is done by way of an atrous spatial pyramid pooling (ASPP)~\citep{deeplabv3} module, which captures multi-scale contextual information using dilated convolutions with different dilation rates. For both training and testing, we used an output stride of 8.

\subsection{Boundary stream}

The \textbf{boundary stream} works essentially as a boundary-detector in our proposed model. It takes as input the same image $I$ as the \textbf{non-boundary stream}, and output a segmentation map highlighting object boundaries as depicted in \Figref{fig:y_model}.

\paragraph{\textbf{Boundary detection}} \label{sec:method-edge-detection} We extract features from different stages of the backbone to improve the prediction accuracy on the boundaries. Features from five different stages, denoted as $F_1$, $F_2$, $F_4$, $F_5$, feed the \textbf{semantic boundary detection} module, as depicted in \Figref{fig:model_complete}. The \textbf{boundary detection} module progressively combines the features $F_1$, $F_3$, $F_4,$ $F_5$ through \textbf{gates} (see the boundary detection module in \Figref{fig:model_complete}) in order to output a set of features $S_n \in \{S_1, S_2, S_3\}$ representing the different processing stages on the boundary that go to the \textbf{semantic boundary detection} module. The use of these gates are necessary to facilitate the flow of information from the \textbf{non-boundary stream} to the \textbf{boundary stream}. Each gate selects from the input features the information that is relevant to perform boundary detection, thus effectively separating the relevant activations to the segmentation from those relevant to the boundary detection tasks. This local gate consists of an attention map $\alpha_g$ followed by a residual basic module. The attention map, $\alpha_g$, is given by
\begin{align}
    \alpha_g &= \sigma(\circledast_{1\times1}(F_n || {F'}_n)) \, , \label{eqn:alpha_gate}
\end{align}
where $F_n$ represents the backbone features from the $n_{th}$ stage used by the boundary detection module, ${\mathbf{F'}}_n$ represents the pre-processed features in the boundary detection module, $||$ is the concatenation function, $\circledast_{1\times1}$ is a convolutional layer, and $\sigma$ is the sigmoid function. 

The gate, $G_A$, is applied to $F_n$, and denoted as
\begin{align}
    \mathrm{G_A} &= \circledast_{1\times1}(F_n \odot ({\alpha_g} + k)) \, , \label{eqn:alpha_gate3}
\end{align}
where $\odot$ is an element-wise product. The resulting $\mathrm{G_A}$ is passed to a residual module, followed by a ${1 \times 1}$ convolution. The final output of each gate is an $S_n$ map. k is a constant to reinforce the borders avoiding close-to-zero values on boundary elements.

\begin{figure*}[t]
    \centering
    \includegraphics[width=0.8\textwidth]{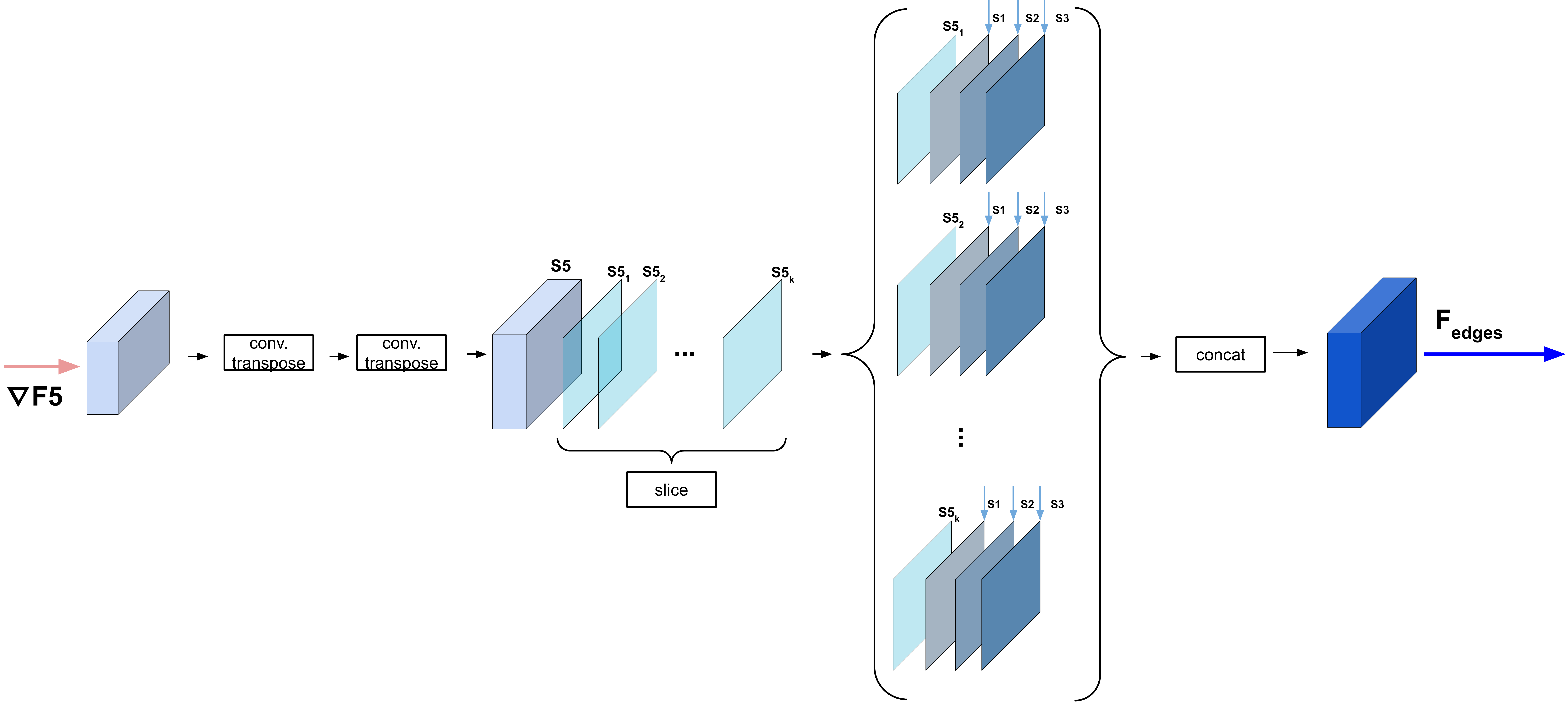}
    \caption{Fusion boundary module. The module takes as input $\nabla F_5$ and $S_n$.
    $\nabla F_5$ is resized by 2 convolution transpose layers. The result is then sliced into $S5_k$ feature maps, with each map being grouped by an $S_n$ set. Finally, an $S5_k$ and $S_n$ are concatenated and passed through a $\circledast_{1\times1}$ layer. This module output is a feature map $\mathrm{F_{edges}} \in \mathbb{R}^{H \times W \times 4 \times L}$, where $\mathrm{L}$ represents the number of classes.}
    \label{fig:fusion_strategy}
\end{figure*}

\paragraph{\textbf{Semantic boundary detection}} \label{sec:method-semantic-edge-detection} This is inspired by the work found in \citep{dff}, where the processing of low-level feature fusion is improved by more semantic features. Unlike \citep{dff}, however, we used the $\nabla F_5$ as an input, in addition to the set of low level features $S_n$ in the \textbf{fusion boundary module}. In our fusion model, differently from what is done by the CASENet~\citep{casenet} network, first we weigh the feature maps from different levels of the network to then perform the ${1 \times 1}$ convolution that ultimately combines the features from multiple stages. $\nabla F_5$ is defined as an approximation of the gradient, and is given by
\begin{align}
    \nabla F_5 &\approx \sigma(F_5 - MaxPool_{3\times3}(F_5)) \label{eqn:sbd_nabla},
\end{align}
where $MaxPool_{3\times3}$ represents a maximum-pooling operation in the 2D space with stride of 3 in both dimensions.  The use of $\nabla F_5$ allows to obtain the most prominent features from $F_5$, which improves the flow of information during back-propagation.

The \textbf{SBD} module is divided into two components: the \textbf{fusion boundary module} and the attention step. The \textbf{fusion boundary module} is shown in \Figref{fig:fusion_strategy}, and takes as inputs $\nabla F_5$ and $S_n$. First, $\nabla F_5$ is upsampled by two convolution transpose layers with stride of 8 pixels. The result is then split into $k$ slices. Each slice, $S5_k$, is concatenated with the $S_n$ sets and linearly combined by a $\circledast_{1\times1}$. The fusion boundary module outputs the $F_{edges}$, combining the sequential application of two $\circledast_{1\times1}$ with batch normalization and a ReLU activation function, $L_{adapt}$, through a semantic boundary attention gate, $\mathrm{SBatt}$, given by
\begin{align}
    \mathrm{SBatt} &= L_{adapt}(\nabla F_5) \odot F_{edges} \label{eqn:alpha_gate2},
\end{align}
The rationale here is to reinforce the boundaries formed by both low-level and high-level features with more semantic information. Finally, the resulting $\mathrm{SBatt}$ is convoluted by a $\circledast_{1\times1}$, resulting in the semantic boundary, $\mathbf{SB} \in \mathbb{R}^{H \times W \times L}$.

\subsection{Semantic fusion gate} \label{sec:method-attention-model}

\begin{figure*}[t]
    \centering
    \includegraphics[width=0.8\textwidth]{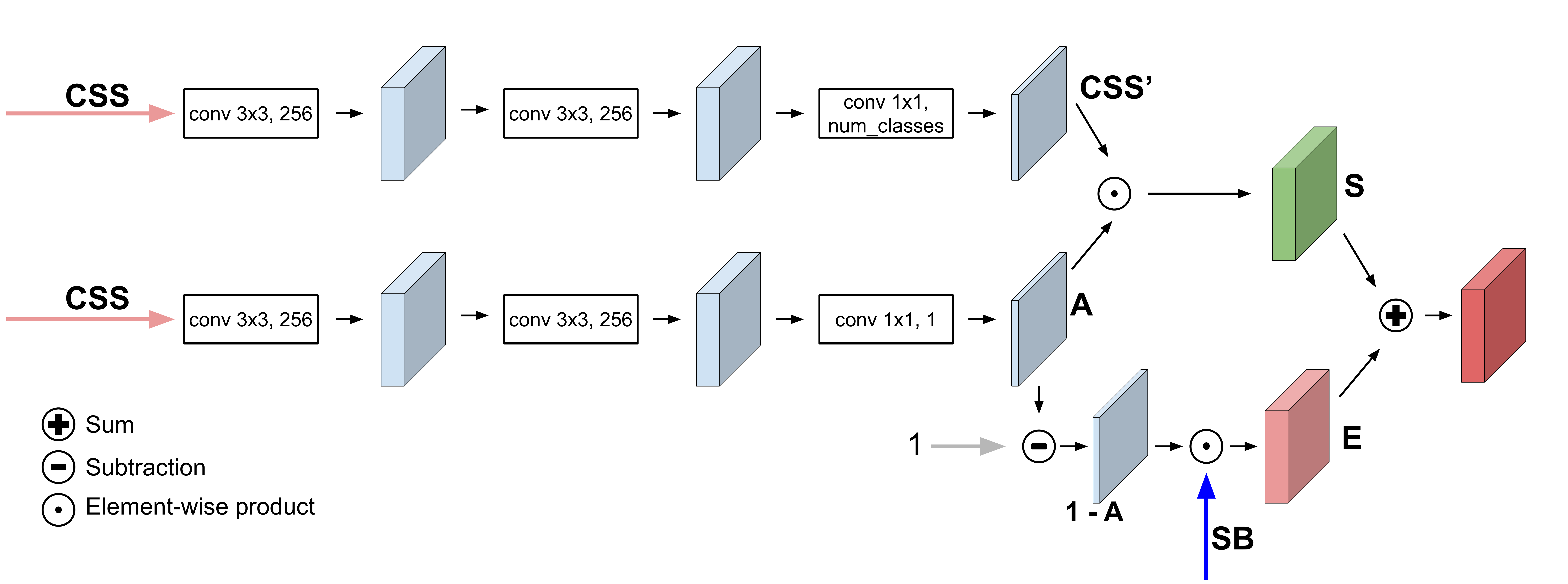}
    \caption{Details of the semantic fusion gate. The module takes as input $\mathrm{CSS}$ features, which are used in two branches. A attention gate model is applied by generating two spatial maps of attention $\mathrm{A}$ and $\mathrm{1-A}$. These maps are used to highlight the features in $\mathrm{CSS'}$ and $\mathrm{SE}$.}
    \label{fig:attenttion_on_edges}
\end{figure*}

\Figref{fig:attenttion_on_edges} illustrates the architecture of the semantic fusion gate, proposed in this work to separate the non-boundary and boundary information. Given a $\mathrm{CSS} \in \mathbb{R}^{H \times W \times L}$ and a semantic boundary $\mathrm{SB} \in \mathbb{R}^{H \times W \times L}$, we first feed $\mathrm{CSS}$ into 2 convolutional sets. The output of the first set is denoted as $\mathrm{CSS'} \in \mathbb{R}^{H \times W \times L}$ features, while the second set outputs a mask, $\mathrm{A} \in \mathbb{R}^{H \times W \times L}$, that works as an attention gate. A $\mathrm{softmax}$ function is multiplied by $\mathrm{CSS'}$, obtaining the segmentation, $\mathrm{S}$. The input $\mathrm{SB}$ is multiplied by $1 - \mathrm{A}$ to get the output $\mathrm{E}$. The final segmentation is finally obtained by $\mathrm{S} + \mathrm{E}$.

The semantic fusion gate shown in \Figref{fig:attenttion_on_edges} split the semantic segmentation features into two distinct tasks. One that represents the pixels with a semantic border and the other with the pixels outside that border. This is so to avoid ambiguity in the classification of border pixels.

\subsection{Multi-task loss}

To train our Y-model, we used different loss functions for each sub-task. We used a weighted binary cross-entropy loss (WBCE) for the boundary stream, and a binary cross-entropy (BCE) loss for the semantic segmentation (non-boundary) stream. Our multi-task loss is modeled as
\begin{align}
\label{eqn_multi_task_loss}
\mathcal{L} &= \lambda_1 \mathcal{L}_{WBCE}(s, \hat{s})+\lambda_2 \mathcal{L}_{BCE}(y,\hat{y}) \, ,
\end{align}
where $\hat{s} \in \mathbb{R}^{H \times W}$ denotes the ground truth of the semantic boundaries, while $\hat{y} \in \mathbb{R}^{H \times W}$ denotes the ground truth of the semantic segmentation. $s$ and $y$ represent the predicted values of the boundary and segmentation, respectively. $\lambda_1$ and $\lambda_2$ are two hyper-parameters that control the weighting between the losses.

To compare the results of the proposed model with other state-of-the-art methods in the Cityscapes and Mapillary Vistas data sets, we used the online hard-example mining (OHEM)~\citep{shrivastava2016training} loss instead of BCE, since this loss function is the one used by the other methods. 

\section{Materials and methods}

\subsection{Training data}
We conducted our experiments in four publicly available data sets: Cityscapes~\citep{Cordts2016Cityscapes}, Mapillary Vistas~\citep{MVD2017}, CamVid~\citep{BrostowSFC:ECCV08_camvid}, and Pascal Context~\citep{mottaghi_cvpr14_pascal_context}. These data sets were chosen by considering the need for annotations with fine granularity and well-defined boundaries, so that the training of the boundary-detection portion of our Y-model was carried out properly. None of these data sets has annotations for semantic boundaries, demanding us to derive the ground truth of the boundaries from the semantic segmentation annotations. We used the distance transform to select pixels, which belong to the boundaries of each class of the semantic segmentation annotations. % Below the main characteristics of each data set are presented.

\subsection{Data augmentation}
As data augmentation strategies, we applied random flips, random scaling in a range of 0.5-2, and crops with a fixed size. We also used photometric distortions such as Gaussian blur and variations to the brightness, hue, and saturation.

% \begin{figure*}[t]
%     \centering
%     \includegraphics[width=0.8\textwidth]{figs/uniform.pdf}
%     \caption{Distribution of the pixels per class in the Cityscapes train set.}
%     \label{fig:pixels_distribution_in_cityscapes}
% \end{figure*}

\begin{table*}[t]
\centering
% \cmidrule{2-23}
\setlength{\tabcolsep}{4pt}
\caption{Comparison between random, inform, integral crop augmentation strategies to train DeeplabV3 and DeeplabV3+ in Cityscapes validation set.}
 \begin{adjustbox}{max width=\textwidth}
\begin{tabular}{lll|ccccccccccccccccccc|r}
\toprule
              method &  backbone        &  crop   &         \rotatebox{90}{road}  &  \rotatebox{90}{s.walk} &        \rotatebox{90}{build.} &        \rotatebox{90}{wall} &    \rotatebox{90}{fence} &          \rotatebox{90}{pole} &       \rotatebox{90}{t-light} &        \rotatebox{90}{t-sign} &           \rotatebox{90}{veg} &       \rotatebox{90}{terrain} &           \rotatebox{90}{sky} &        \rotatebox{90}{person} &         \rotatebox{90}{rider} &           \rotatebox{90}{car} &         \rotatebox{90}{truck} &           \rotatebox{90}{bus} &         \rotatebox{90}{train} &         \rotatebox{90}{motor} &       \rotatebox{90}{bicycle} &          mean \\
% method & backbone & crop &               &               &               &               &               &               &               &               &               &               &               &               &               &               &               &               &               &               &               &               \\
\midrule
\midrule

DeeplabV3 & resnet50 & random &          97.7 &           82.6 &           91.3 & \textbf{ 49.3} &           58.6 & \textbf{ 52.4} &           65.3 &           74.1 &           91.0 &           58.0 & \textbf{ 93.4} &           77.2 &           55.5 &           93.9 &           71.7 &           82.5 &           67.0 &           59.9 & \textbf{ 73.9} &           73.4 \\
          &          & uniform &           96.1 &           75.5 &           90.1 &           37.4 &           50.6 &           52.2 &           63.3 &           74.6 &           89.4 &           40.8 &           92.2 &           74.6 &           52.8 &           91.9 &           55.5 &           83.7 &           68.0 & \textbf{ 61.4} &           73.5 &           69.7 \\
          &          & integral & \textbf{ 97.9} & \textbf{ 83.3} & \textbf{ 91.3} &           47.5 & \textbf{ 60.1} &           52.2 & \textbf{ 65.5} & \textbf{ 74.7} & \textbf{ 91.0} & \textbf{ 59.7} &           93.3 & \textbf{ 77.8} & \textbf{ 56.8} & \textbf{ 94.1} & \textbf{ 80.5} & \textbf{ 84.6} & \textbf{ 71.9} &           61.2 &           73.7 & \textbf{ 74.6} \\ \hline
              & resnet101 & random &                97.9 &           82.7 &           91.5 &           55.0 &           59.0 &           52.7 &           66.3 &           74.5 &           91.0 &           48.9 & \textbf{ 93.4} &           77.7 &           56.7 &           94.1 & \textbf{ 79.0} & \textbf{ 86.9} &           63.5 & \textbf{ 65.0} &           74.9 &           74.3 \\
          &          & uniform &           97.3 &           78.5 &           90.7 &           38.9 &           54.5 &           53.0 &           64.3 &           70.9 &           89.8 &           35.7 &           92.5 &           76.3 &           57.5 &           93.5 &           76.5 &           80.9 &           63.5 &           60.5 &           74.1 &           71.0 \\
          &          & integral & \textbf{ 98.1} & \textbf{ 84.1} & \textbf{ 91.7} & \textbf{ 55.7} & \textbf{ 60.3} & \textbf{ 53.1} & \textbf{ 67.3} & \textbf{ 75.2} & \textbf{ 91.2} & \textbf{ 54.0} &           93.3 & \textbf{ 78.2} & \textbf{ 57.9} & \textbf{ 94.2} &           77.0 &           83.5 & \textbf{ 72.0} &           62.7 & \textbf{ 74.9} & \textbf{ 75.0}  \\ \hline
DeeplabV3+ & resnet50 & random & \textbf{ 98.0} & \textbf{ 84.5} & \textbf{ 92.3} & \textbf{ 48.2} &           59.3 &           65.5 & \textbf{ 70.4} &           78.9 &           92.1 &           59.2 & \textbf{ 94.9} &           81.3 &           58.9 &           95.1 &           73.7 &           83.2 &           71.0 &           59.5 &           76.5 &           75.9 \\
          &          & uniform &           97.4 &           81.3 &           91.6 &           39.4 &           55.2 &           65.3 &           67.3 &           78.1 &           90.9 &           47.8 &           94.0 &           80.3 &           58.7 &           94.0 &           70.5 &           83.8 &           67.3 & \textbf{ 60.8} &           76.4 &           73.7 \\
          &          & integral &           97.9 &           84.3 &           92.2 &           45.1 & \textbf{ 59.5} & \textbf{ 65.8} &           69.9 & \textbf{ 79.1} & \textbf{ 92.1} & \textbf{ 60.5} &           94.4 & \textbf{ 81.6} & \textbf{ 59.5} & \textbf{ 95.1} & \textbf{ 75.7} & \textbf{ 85.7} & \textbf{ 71.8} &           59.3 & \textbf{ 76.7} & \textbf{ 76.1} \\ \hline
              & resnet101 & random &           98.2 &           84.9 &           92.5 &           46.8 &           59.6 & \textbf{ 66.9} &           70.4 & \textbf{ 79.9} &           91.9 &           52.5 & \textbf{ 94.8} &           81.6 &           59.8 & \textbf{ 95.3} & \textbf{ 77.4} &           83.8 &           71.3 & \textbf{ 63.3} & \textbf{ 77.3} &           76.2 \\
          &          & uniform &           97.6 &           81.2 &           91.9 &           40.0 &           57.1 &           66.7 &           68.5 &           79.3 &           91.2 &           47.0 &           94.1 &           80.2 &           58.8 &           94.2 &           76.6 & \textbf{ 85.1} & \textbf{ 80.7} &           61.4 &           76.9 &           75.2 \\
          &          & integral & \textbf{ 98.2} & \textbf{ 85.2} & \textbf{ 92.7} & \textbf{ 49.0} & \textbf{ 61.3} &           66.9 & \textbf{ 71.4} &           79.7 & \textbf{ 92.3} & \textbf{ 55.8} &           94.8 & \textbf{ 82.0} & \textbf{ 60.8} &           95.1 &           76.9 &           84.8 &           77.8 &           61.5 &           77.1 & \textbf{ 77.0} \\
\bottomrule
\end{tabular} 
\end{adjustbox}
\label{tab:comp_baseline_cityscapes_integral}

\end{table*}

%{\color{red} To depict an example of how unbalanced a data set can be, \Figref{fig:pixels_distribution_in_cityscapes} presents the distribution of the number of pixels per class on Cityscapes train set. In this data set, the number of pixels of the road and car classes is greater than those of the train and bicycle classes.}

\paragraph{Integral crop} Class imbalance occurs when one class is significantly more prevalent than others in the training data set. This is a common problem present in all data sets used. This distribution imbalance between classes impairs the training of segmentation models, particularly making it difficult for them to learn less present classes. In general, a model trained with an unbalanced data set achieves lower performance and has less generalization capability than a model trained on a uniform data set. Our proposed cropping method takes each pixel in the image into account to identify the optimal crop to be used during training. This is done by assigning an individual weight value to each pixel. We find the crop region that maximizes the sum value of the weights for each pixel in the region, and consider that as the best crop. By considering the maximum value, we ensure that rarer classes have a greater chance of appearing in the crop. Calculating the sum of each possible crop region can be very computationally expensive. To do this efficiently, we explored the computation of integral images. The integral image algorithm is capable of calculating the sum of values in a rectangular subset of a grid in an efficient manner, and constant time. To assess whether this strategy had a positive impact on the training of a segmentation model, we carried out comparative evaluations on Cityscapes validation data set whose results are laid out in Table \ref{tab:comp_baseline_cityscapes_integral}. To perform these experimental analysis (and also the ablation study in Section \ref{ablation}) two networks were used as baselines: DeeplabV3 and DeeplabV3+. These networks were chosen to motivate our gains over strong baselines. It is noteworthy that integral crop achieved the best results in comparison with random and uniform crops (see \textbf{mean} column); henceforth, we follow with our proposed crop in the ablation studies and comparative analysis when using our proposed models.

\subsection{Implementation details}

%Code for our proposed model was written by using Tensorflow. As in previous works \citep{yu2015multi, zhang2018context}, 
We used a Dilated ResNet \citep{dilated-resnet} backbone, which was pre-trained on ImageNet 1k \citep{imagenet_cvpr09}, with dilation in the last two stages and output size of 1/8. We followed the work in \citep{pspnet, deeplabv3} considering a learning rate schedule as $lr=baselr * (1-\frac{iter}{total\_iter})^{power}$, equal to 0.01 and power of 0.9. The SGD optimizer was parameterized with a momentum of 0.9 and weight decay set to 0.0005. All the training stage was done on a V100 NVIDIA DGX Station with 8 GPUs. The batch size was 8 for Cityscapes, Mapillary Vistas and CamVid, and 16 for Pascal Context. We used fixed crop with values equals to $800 \times 800$ for Cityscapes, Mapillary Vistas, and CamVid. For Pascal Context, we used $512 \times 512$ as a fixed size for the crop. To increase training speed, we used mixed-precision training \citep{narang2018mixed}. To improve the results, we used sync cross-gpu batch normalization following the works in \citep{zhang2018context, dualattention1, dualattention2}.

\subsection{Evaluation Metric}
For evaluation, we used the standard evaluation metric of mean intersection over union (mIoU). In the Cityscapes, CamVid, and Mapillary Vistas data sets, backgrounds are disregarded from the results. On the Pascal Context data sets, however, we used the official evaluation, which considers the background as a class in the mIoU calculation. To evaluate whether the proposed model improves when we only consider the boundaries, we used the f1-boundary metric \citep{perazzi2016benchmark}.

\section{Experimental evaluation} \label{experiment}

To assess the full performance of our proposed model, we first analyzed the influence of using the boundary stream and the semantic fusion gate model compared with a baseline model. For that, an ablative study was carried out on the Cityscapes validation set. Yet the effectiveness of the boundary detection was evaluated by varying the border thickness of the image boundaries. Finally, we compared our best setup with twelve other state-of-the-art methods in the literature. 

%TODO explicar que a concatenação foi feita por concatenação.
\begin{table}[t] 
\centering
\caption{Results on Cityscapes \textbf{validation set} with different configurations of our Y-model.}
\resizebox{1\linewidth}{!}
 {
% \rowcolors{2}{}{lightgray}
\newcolumntype{g}{>{\columncolor{lightgray}}c}
 \begin{tabular}{l | g g | g}
 \rowcolor{white}
 \hline  {\bf Baseline} & {\bf boundary stream} &  {\bf sf gate} & \bf{mIoU\%} \\ [0.5ex]
 \hline\hline
 	\multirow{4}{*}{ DeeplabV3 (ResNet101)} &  &  & 77.2 \\
    & {\cellcolor{white} \checkmark} & {\cellcolor{white}}  &{\cellcolor{white}73.6} \\
    & & \checkmark & 76.5 \\
    & {\cellcolor{white}\checkmark} & {\cellcolor{white}\checkmark} & {\cellcolor{white} \bf 78.2} \\ \hline
     \multirow{4}{*}{DeeplabV3+ (ResNet101) } &  &  & 77.0 \\
    & {\cellcolor{white}\checkmark} & {\cellcolor{white}} & {\cellcolor{white}75.2} \\
    & & \checkmark & 76.1 \\
    \rowcolor{white}
    & \checkmark & \checkmark & {\bf 78.9} \\ \hline
 \end{tabular}  }
 \label{tab:comp_different_type_sbcnn}
\end{table}

\begin{table*}[t]
    \setlength{\tabcolsep}{1pt}
    \tiny
    \newcommand{\C}{0.06\textwidth}
    \renewcommand{\arraystretch}{1.5}% Spread rows out..
    \centering
    \caption{Per-class results in the Cityscapes validation set our Y-model. All networks were training considering ResNet101 as the backbone.}
    
    \resizebox{\textwidth}{!}{
    \begin{tabular}{l|ccccccccccccccccccc|c}
\hline
Method &          \rotatebox{90}{road}  &  \rotatebox{90}{s.walk} &        \rotatebox{90}{build.} &        \rotatebox{90}{wall} &    \rotatebox{90}{fence} &          \rotatebox{90}{pole} &       \rotatebox{90}{t-light} &        \rotatebox{90}{t-sign} &           \rotatebox{90}{veg} &       \rotatebox{90}{terrain} &           \rotatebox{90}{sky} &        \rotatebox{90}{person} &         \rotatebox{90}{rider} &           \rotatebox{90}{car} &         \rotatebox{90}{truck} &           \rotatebox{90}{bus} &         \rotatebox{90}{train} &         \rotatebox{90}{motor} &       \rotatebox{90}{bicycle} &          mean
 \\
\hline
\hline
DeepLabV3 \citep{deeplabv3} &         98.0 &       83.9 &       91.1 &       52.8 &       59.4 &       49.4 &          64.8 &         72.1 &       91.1 &       64.2 &       92.7 &       77.8 &       60.1 &       93.9 &       79.3 &       85.5 &       71.1 &       68.0 &       74.6 &       75.3 \\
DeepLabV3+ \citep{deeplabv3plus} &       98.2 &       85.2 &       92.7 &       49.0 &       61.3 &       66.9 &          71.4 &         79.7 &       92.3 &       55.8 &       94.8 &       82.0 &       60.8 &       95.1 &       76.9 &       84.8 &       77.8 &       61.5 &       77.1 &       77.0 \\
Ours (baseline   DeepLabV3) &       98.0 &       85.0 &       91.0 & {\bf 53.0} &       61.0 &       64.0 &          72.0 &         79.0 &       92.0 & {\bf 65.0} &       93.0 &       83.0 & {\bf 64.0} &       95.0 &       73.0 &       87.0 &       70.0 & {\bf 72.0} &       78.0 &       78.2 \\
Ours  (baseline DeepLabV3+) &  {\bf 98.4} & {\bf 86.4} & {\bf 93.1} &       51.5 & {\bf 64.9} & {\bf 68.8} &    {\bf 73.6} &   {\bf 81.4} & {\bf 92.7} &       58.9 & {\bf 95.0} & {\bf 83.4} &       63.9 & {\bf 95.5} & {\bf 80.9} & {\bf 87.8} & {\bf 80.9} & 64.2 & {\bf 78.1} & {\bf 78.9} \\
\hline
    \end{tabular}
    }
    \label{tab:comp_baseline_integral}
\end{table*}

\subsection{Ablation study} \label{ablation}

We hypothesize that improved boundary detection leads to a better overall segmentation. To evaluate this, we assessed the performance of the additional main modules of our Y-model in comparison with a baseline: The boundary stream and the semantic fusion gate (attention module). For that, we first considered a truth table with the four possible setups with the main modules, DeeplabV3 and DeeplabV3+ as baselines, and ResNet101 as backbone. The results are summarized in Table \ref{tab:comp_different_type_sbcnn}. Our findings showed that the use of both modules greatly improved both baselines on the Cityscapes validation set, having the best results when using DeeplabV3+. It is noteworthy that when employing only the boundary stream, the results are 1.7 and 1.8 percentage points worse than DeeplabV3 and DeeplabV3+, respectively. This lead us to conclude that the performance of semantic segmentation suffers when using only the boundary stream without our proposed fusion model. Considering only the attention module, there is a gain of 1.2 percentage points over DeeplabV3, while a decrease of 0.9 percentage point over DeeplabV3+. Fusing both modules can significantly improve the performance of the final semantic segmentation.

Table \ref{tab:comp_baseline_integral} derives from Table \ref{tab:comp_different_type_sbcnn} and show the results found by class. Our Y-model, when based on DeepLabV3+, outperforms the two baseline models for all classes but wall, terrain, rider, and motor. The final gain over DeeplabV3 and DeeplabV3+ was 2.9 and 1.9 percentage points, respectively.

\subsection{Evaluating semantic boundary detection}

To check if our method improves the boundary detection, we evaluated the results of the f1-boundary score in the Cityscapes validation set. The results are summarized in Table \ref{tab:comp_baseline_cityscapes_f1}. The f1-boundary score represents the contour matching between the predicted segmentation and the ground truth. By using the f1-boundary score, we can determine the thickness of the boundary (\textbf{B. Thick}) that we will consider in the evaluation. This is so to allow us to assess whether our model produces good boundary detection results with respect to the baselines DeepLabV3 and DeepLabV3+. For the evaluation, we considered four values of boundary thickness: 3, 5, 9, and 12. When considering the \textbf{mean}, our Y-model performed the best for all values of thickness using both baselines. These findings show that our Y-model has superior results in terms of boundary detection even in comparison with strong baselines such as DeepLabV3 and DeepLabV3+.

\begin{table*}[t]
% \cmidrule{2-23}
\setlength{\tabcolsep}{4pt}
\caption{Comparison between the baselines and our Y-model in the Cityscapes validation set, considering different thresholds in f1-boundary score. }  
\begin{adjustbox}{max width=\textwidth}
\begin{tabular}{lr|rrrrrrrrrrrrrrrrrrr|r}
    \toprule
B. Thick. &  Method & \rotatebox{90}{road}  &  \rotatebox{90}{s.walk} &        \rotatebox{90}{build.} &        \rotatebox{90}{wall} &    \rotatebox{90}{fence} &          \rotatebox{90}{pole} &       \rotatebox{90}{t-light} &        \rotatebox{90}{t-sign} &           \rotatebox{90}{veg} &       \rotatebox{90}{terrain} &           \rotatebox{90}{sky} &        \rotatebox{90}{person} &         \rotatebox{90}{rider} &           \rotatebox{90}{car} &         \rotatebox{90}{truck} &           \rotatebox{90}{bus} &         \rotatebox{90}{train} &         \rotatebox{90}{motor} &       \rotatebox{90}{bicycle} &    mean\\
\midrule
\multirow{4}{*}{3px} & DeepLabV3                 &       80.8 &       57.6 &       60.9 &       48.5 & {\bf 48.1} &       53.8 &          53.2 &         57.0 &       61.8 &       50.2 &       73.0 &       50.8 &       61.9 &       70.4 & {\bf 74.1} &       84.0 &       90.0 & {\bf 74.8} &       55.0 &       63.4 \\
 & Ours (baseline DeepLabV3) & {\bf 83.7} & {\bf 64.8} & {\bf 70.2} & {\bf 53.7} &       46.0 & {\bf 73.3} &    {\bf 62.0} &   {\bf 69.6} & {\bf 71.6} & {\bf 53.6} & {\bf 81.7} & {\bf 63.9} & {\bf 62.5} & {\bf 79.6} &       73.8 & {\bf 88.1} & {\bf 93.6} &       72.2 & {\bf 60.5} & {\bf 69.7} \\
    \cmidrule{2-22}
& DeepLabV3+                  &       83.7 &       64.8 & {\bf 70.2} &       53.7 &       46.0 &       73.3 &          62.0 &         69.6 &       71.6 &       53.6 & {\bf 81.7} &       63.9 &       62.5 &       79.6 &       73.8 & {\bf 88.1} &       93.6 &       72.2 &       60.5 &       69.7 \\
& Ours (baseline DeepLabV3+)  & {\bf 83.8} & {\bf 64.9} &       69.5 & {\bf 56.2} & {\bf 50.4} & {\bf 74.0} &    {\bf 73.4} &   {\bf 74.1} & {\bf 71.7} & {\bf 53.8} &       81.2 & {\bf 67.8} & {\bf 69.7} & {\bf 80.8} & {\bf 81.4} &       87.7 & {\bf 93.6} & {\bf 78.9} & {\bf 64.1} & {\bf 72.5} \\
\cmidrule{2-22}
%% & GSCNN (baseline WideResNet)* & {\bf 85.0} & {\bf 68.8} & {\bf 74.1} & {\bf 53.3} &       47.0 & {\bf 79.6} &    {\bf 74.3} &   {\bf 76.2} & {\bf 75.3} & {\bf 53.1} & {\bf 83.5} & {\bf 69.8} & {\bf 73.1} & {\bf 83.4} &       75.8 & {\bf 88.0} & {\bf 93.9} &       75.1 & {\bf 68.5} & {\bf 73.6} \\
    \midrule
    \midrule
\multirow{4}{*}{5px} & DeepLabV3 &  \textbf{86.2} &          69.0 & \textbf{73.5} &          51.6 &          51.6 &          69.0 &          61.7 & \textbf{70.5} &          75.6 &          54.8 & \textbf{82.6} &          63.2 &          68.5 & \textbf{82.2} &          75.4 &          85.5 &          90.3 &          76.7 & \textbf{64.6}  &          71.2 \\
  & Ours (baseline DeepLabV3)      &          86.0 & \textbf{69.0} &          73.2 & \textbf{52.3} & \textbf{53.1} & \textbf{69.0} & \textbf{66.6} &          69.4 & \textbf{75.7} & \textbf{55.5} &          81.5 & \textbf{63.5} & \textbf{69.3} &          82.1 & \textbf{78.4} & \textbf{88.1} & \textbf{92.1} & \textbf{79.3} &          63.9  & \textbf{72.0} \\
    \cmidrule{2-22}
& DeepLabV3+                 & {\bf 88.0} &       71.9 & {\bf 77.9} &       56.2 &       49.0 &       78.2 &          67.8 &         75.7 &       80.4 &       57.4 & {\bf 86.8} &       70.5 &       67.5 &       85.5 &       74.8 & {\bf 89.3} &       93.8 &       73.8 &       67.5 &       74.3 \\
& Ours (baseline DeepLabV3+) &       87.9 & {\bf 72.2} &       76.9 & {\bf 58.8} & {\bf 53.4} & {\bf 78.5} &    {\bf 79.0} &   {\bf 79.8} & {\bf 80.6} & {\bf 57.7} &       86.6 & {\bf 74.4} & {\bf 74.8} & {\bf 86.7} & {\bf 82.6} &       88.8 & {\bf 93.9} & {\bf 80.4} & {\bf 71.0} & {\bf 77.1} \\
\cmidrule{2-22}
% & GSCNN (baseline WideResNet)* & {\bf 88.7} & {\bf 75.3} & {\bf 80.9} & {\bf 55.9} &       49.9 & {\bf 83.6} &    {\bf 78.6} &   {\bf 80.4} & {\bf 83.4} & {\bf 56.6} & {\bf 88.4} & {\bf 75.4} & {\bf 77.8} & {\bf 88.3} &       77.0 & {\bf 88.9} & {\bf 94.2} &       76.9 & {\bf 75.1} & {\bf 77.6} \\
    \midrule
    \midrule
 \multirow{4}{*}{9px} & DeepLabV3  &          90.3 &          76.5 &          82.9 &          54.8 &          55.2 &          77.6 &          68.1 & \textbf{78.4} &          85.7 &          58.9 & \textbf{88.0} &          72.2 &          74.8 & \textbf{89.3} &          76.8 &          86.9 &          90.7 &          78.6 & \textbf{73.4}  &          76.8 \\
  & Ours (baseline DeepLabV3) & \textbf{90.4} & \textbf{76.8} & \textbf{82.9} & \textbf{55.6} & \textbf{57.0} & \textbf{77.9} & \textbf{72.9} &          78.0 & \textbf{86.3} & \textbf{59.6} &          87.6 & \textbf{72.8} & \textbf{74.9} &          89.3 & \textbf{79.6} & \textbf{89.5} & \textbf{92.6} & \textbf{81.1} &          73.1 & \textbf{77.8} \\
    \cmidrule{2-22}
 & DeepLabV3+                 &       90.9 &       77.2 & {\bf 84.1} &       58.9 &       52.1 &       81.5 &          71.9 &         79.5 &       86.9 &       60.8 &       89.6 &       75.5 &       72.0 &       89.6 &       75.9 & {\bf 90.3} & {\bf 94.2} &       75.1 &       73.7 &       77.9 \\
 & Ours (baseline DeepLabV3+) & {\bf 90.9} & {\bf 77.6} &       82.8 & {\bf 61.6} & {\bf 56.6} & {\bf 81.7} &    {\bf 82.9} &   {\bf 83.5} & {\bf 87.3} & {\bf 61.5} & {\bf 89.6} & {\bf 79.2} & {\bf 79.2} & {\bf 90.8} & {\bf 83.9} &       89.7 &       94.1 & {\bf 82.0} & {\bf 77.3} & {\bf 80.6} \\
 \cmidrule{2-22}
% & GSCNN (baseline WideResNet)* & {\bf 91.3} & {\bf 80.1} & {\bf 86.0} & {\bf 58.5} &       52.9 & {\bf 86.1} &    {\bf 81.5} &   {\bf 83.3} & {\bf 89.0} & {\bf 59.8} & {\bf 91.1} & {\bf 79.1} & {\bf 81.5} & {\bf 91.5} &       78.1 & {\bf 89.7} & {\bf 94.4} &       78.5 & {\bf 80.4} & {\bf 80.7} \\
    \midrule
    \midrule
\multirow{4}{*}{12px} & DeepLabV3 &          91.5 &          78.7 &          85.7 &          56.2 &          56.6 &          79.4 &          69.8 & \textbf{80.2} &          88.5 &          60.3 & \textbf{89.3} &          74.5 & \textbf{76.9} & \textbf{91.1} &          77.4 &          87.4 &          90.9 &          79.3 & \textbf{76.5} &          78.4 \\
 & Ours (baseline DeepLabV3) &\textbf{91.6} & \textbf{79.1} & \textbf{85.8} & \textbf{56.9} & \textbf{58.5} & \textbf{79.8} & \textbf{74.6} &          80.0 & \textbf{89.2} & \textbf{61.2} &          89.0 & \textbf{75.2} &          76.9 &          91.0 & \textbf{80.2} & \textbf{90.0} & \textbf{92.8} & \textbf{81.8} &          76.5 & \textbf{79.5} \\
 \cmidrule{2-22}
& DeepLabV3+                 &       91.9 &       79.1 & {\bf 86.4} &       60.2 &       53.4 &       83.0 &          73.1 &         80.8 &       89.1 &       62.2 & {\bf 90.6} &       77.2 &       73.6 &       91.1 &       76.4 & {\bf 90.7} & {\bf 94.3} &       75.8 &       76.0 &       79.2 \\
& Ours (baseline DeepLabV3+) & {\bf 92.0} & {\bf 79.5} &       85.1 & {\bf 62.8} & {\bf 58.1} & {\bf 83.1} &    {\bf 84.0} &   {\bf 84.8} & {\bf 89.6} & {\bf 63.1} &       90.5 & {\bf 80.9} & {\bf 81.0} & {\bf 92.2} & {\bf 84.3} &       90.0 &       94.2 & {\bf 82.6} & {\bf 79.7} & {\bf 82.0} \\
\cmidrule{2-22}
% & GSCNN (baseline WideResNet)* & {\bf 92.2} & {\bf 81.7} & {\bf 87.9} & {\bf 59.6} &       54.3 & {\bf 87.1} &    {\bf 82.3} &   {\bf 84.4} & {\bf 90.9} & {\bf 61.1} & {\bf 91.9} & {\bf 80.4} & {\bf 82.8} & {\bf 92.6} &       78.5 & {\bf 90.0} & {\bf 94.6} &       79.1 & {\bf 82.2} & {\bf 81.8} \\
    \midrule
\bottomrule
\end{tabular}
\end{adjustbox}
\label{tab:comp_baseline_cityscapes_f1}

\end{table*}

\subsection{Comparison with the state-of-the-art}

We compared our method with twelve other methods \citep{deeplabv3plus,dilated-resnet,peng2017large, pspnet, zhao2018psanet, wu2016high, pspnet, porzi2019seamless, rota2018place, yu2018bisenet, ding2019boundary, yu2015multi}, considered closely related to ours. For the Cityscapes, Mapillary Vistas, and CamVid data sets, we used the WideResNet network with 38 layers \citep{wideresnet} and with expansion for the last 3 residual blocks. We used the extended ResNet network with 101 layers \citep{dilated-resnet} as a backbone for our network on the Pascal Context data sets.

The results found over the selected data sets are presented in Tables \ref{tab:comp_sota_cityscapes}, \ref{tab:comp_sota_mapillary}, and \ref{tab:comp_sota_camvid}, \ref{tab:comp_sota_pascal_context} for the Cityscapes, Mapillary Vistas, CamVid, and Pascal Context data sets, respectively. As we are using the same evaluation protocol that those used in the state-of-the-art methods for all data sets, the results considered in the tables are those reported in each one of their original works. Note also that not all methods are evaluated on all data sets. From what we achieved in the ablation study, our proposed model uses DeeplabV3+ as the backbone of segmentation in all comparisons in this section. Below the results are discussed for each one of the data sets used.

\paragraph{\textbf{Results on Cityscapes}} Table \ref{tab:comp_sota_cityscapes} shows the results of the Cityscapes testing split. We obtained the results with 90k training steps. We did not use coarse annotation because the loss of our Y-model needs fine boundary annotation. Both the training set and the validation set were used for fine-tuning. For prediction, we adopted a multi-scale strategy with values of 0.5, 1, and 2 for all the methods. The best result was achieved by our Y-model with a mIoU of 80.6\%.

\paragraph{\textbf{Results on Mapillary Vistas}} This data set was included to check if our network was able to achieve competitive results on a larger number of classes. To train Mapillary Vistas, we used 110k training steps. We report our results in Mapillary Vistas validation set using a single scale for prediction. As summarized in Table \ref{tab:comp_sota_mapillary}, we achieved the second-best result, being 0.8 percentage point behind the best method.

\paragraph{\textbf{Results on CamVid}} To carry out experiments on CamVid, we first pre-trained our Y-model in the Cityscapes training set. The pre-training with Cityscapes was obtained with 90k training steps. This procedure was necessary due to the CamVid data set provides only 369 images for training. Another important aspect of the images in this data set is that they are sequential, and the training with only these images can cause a model overfit. After pre-training with Cityscapes, only the first layer of each stream was used in the training with CamVid data set. We used 10k training steps to finetuning the training in CamVid. Our Y-model achieved the best result, 1 percentage point better than the second-best method as shown in Table \ref{tab:comp_sota_camvid}, using a single-scale inference. 

\paragraph{\textbf{Results on Pascal Context}}
To evaluate our network in Pascal Context, we used a subset of 59 classes like in \citep{yu2015multi} in a 90k training steps. Table \ref{tab:comp_sota_pascal_context} shows that our network achieved the best result in this data set, 0.3 percentage point better than the second place, using a single-scale inference. 

% \paragraph{\textbf{Results on Pascal VOC 2012}} In this data set, we performed a pre-training with the MS-COCO data set. This pre-training was performed with 60k training steps. For fine-tuning, we used 50k training steps. The results were obtained by using multiple scales with values equal to 0.5, 0.75, 1, 2, as done by \citep{deeplabv3, deeplabv3plus}. According to Table \ref{tab:comp_sota_pascal_voc}, our proposed model reached 82.3\% of mIoU. Our hypothesis for the lowest result achieved is that the scene is not semantically marked by labels that adequately delimit the semantic boundary of the objects. Indeed, this data set has just one or two objects annotated in the scene, and the rest of the image is considered background. This way, the background can represent, in practice, the object fleet with different types of semantic classes, which makes ambiguous the classification of these boundaries. The result in the Pascal Context data set reinforces the hypothesis of why our network presented the worst result in the Pascal VOC 2012 data set, since the former is a complete annotated version of the latter.

\subsection{Qualitative evaluation}

\begin{figure}[t]
    \centering
    % \subfigure[]{\includegraphics[width=.2\textwidth]{figs/attention/frankfurt_000000_001236_leftImg8bit.png}}
    % \subfigure[]{\includegraphics[width=.2\textwidth]{figs/attention/atten2.png}} \\
    \subfigure[]{\includegraphics[width=.24\textwidth]{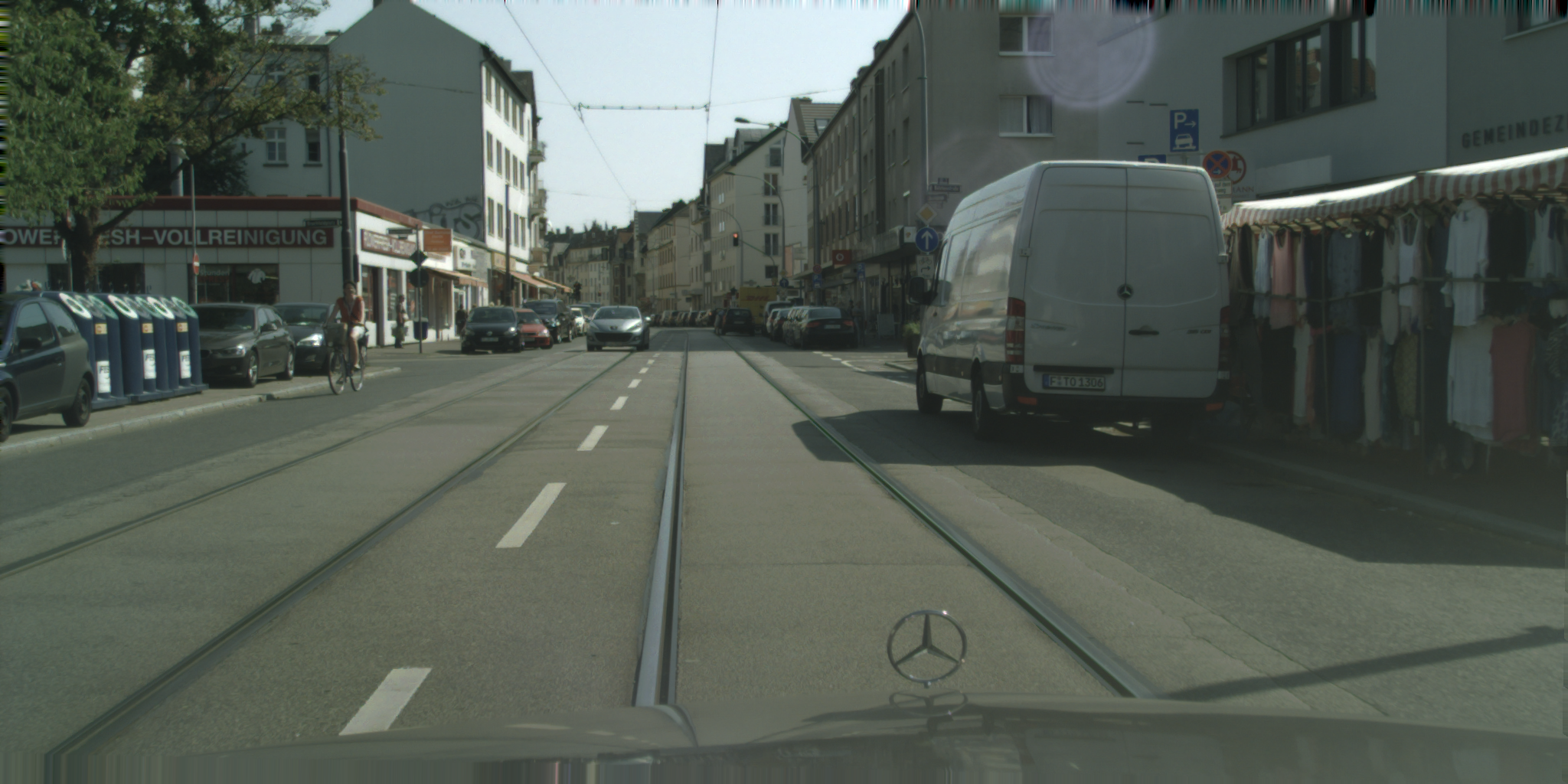}}
    \subfigure[]{\includegraphics[width=.24\textwidth]{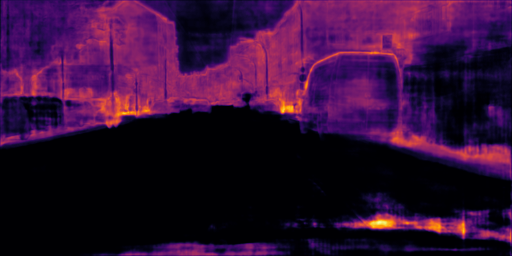}}

    \caption{$\alpha$-maps of the semantic fusion gate depicted as heat maps (b) with respect to the original image (a). The brighter the pixel, the more important it is to the segmentation procedure. Sample image from the Cityscapes validation set.}
    \label{fig:eval_quantitative_results_atten}
\end{figure}

\begin{table}[t]
\centering
\caption{Comparative results in the Cityscapes testing set.}
\resizebox{1\linewidth}{!}
 {
 \begin{tabular}{l|c|c}
 \hline
 {\bf Method } & Multi-scale & {\bf mIoU (\%) } \\ [0.5ex]
 \hline\hline
DeepLabV3+(dilated-ResNet-50)~\citep{deeplabv3plus}  &   & 73.0 \\
Dilated-ResNet-101~\citep{dilated-resnet} &  & 75.7 \\
DeepLabV3+(dilated-ResNet-101)~\citep{deeplabv3plus}  &  & 76.5 \\
Large Kernel Matters~\citep{peng2017large} &   & 76.9 \\
PSPNet(dilated-ResNet-101)~\citep{pspnet}   &  \checkmark & 78.4 \\
PSANet(dilated-ResNet-101)~\citep{zhao2018psanet}   & \checkmark  & 80.1 \\
Our(WideResNet-38)   & \checkmark & {\bf 80.6} \\
 \hline
 \end{tabular}}
 \label{tab:comp_sota_cityscapes}
\end{table}

\begin{table}[t]
\centering
\caption{Comparative results in the Mapillary Vistas validation set.}
% \resizebox{1\linewidth}{!}
 {
 \begin{tabular}{l | c}
 \hline
 {\bf Method } & {\bf mIoU (\%) } \\ [0.5ex] 
 \hline\hline
 FCN(WideResNet-38)~\citep{wu2016high} & 41.1 \\
 FCN(WideResNet-38 + bcs)~\citep{wu2016high}  & 47.7 \\  
 PSPNet(dilated-ResNet101)~\citep{pspnet}  & 49.7 \\ 
 Seamless~\citep{porzi2019seamless} & 50.4 \\ 
 Ours(WideResNet-38) & 52.3 \\ [1.0ex]
 Inplace-ABN~\citep{rota2018place} & {\bf 53.1} \\
 \hline
 \end{tabular}}
 \label{tab:comp_sota_mapillary}
\end{table}

\begin{table}[t]
\centering
\caption{Comparative results in CamVid testing set.}
% \resizebox{1\linewidth}{!}
 {
 \begin{tabular}{l | c}
 \hline
 {\bf Method } & {\bf mIoU (\%) } \\ [0.5ex] 
  \hline 
  \hline
 PSPNet(dilated-ResNet-101)~\citep{pspnet} & 69.1 \\
 BiSetNet~\citep{yu2018bisenet} & 68.7 \\
 Dilated ResNet-101~\citep{dilated-resnet} &  65.3 \\ 
 BFP~\citep{ding2019boundary} & 74.1 \\
 Ours(WideResNet-38) & \textbf{75.1} \\ [1.0ex]
 \hline
 \end{tabular}}
 \label{tab:comp_sota_camvid}
\end{table}

\begin{table}[t]
\centering
\caption{Comparative results in Pascal Context validation set.}
% \resizebox{1\linewidth}{!}
 {
 \begin{tabular}{l | c}
 \hline
 {\bf Method } & {\bf mIoU (\%) } \\ [0.5ex] 
  \hline\hline
 Dilated-ResNet-101~\citep{dilated-resnet} &  42.6 \\ 
 RefineNet~\citep{yu2015multi} & 47.3 \\
 PSPNet(dilated-ResNet101)~\citep{pspnet} & 47.8 \\
 Ours(dilated-ResNet101)  & {\bf48.1} \\ [1.0ex]
 \hline
 \end{tabular}}
 \label{tab:comp_sota_pascal_context}
\end{table}

\paragraph{\textbf{Attention maps}} The semantic fusion gate divides the input image pixels into two groups: Those related to the segmentation of boundary regions and those related to the segmentation of non-boundary regions. \Figref{fig:eval_quantitative_results_atten} shows the $\alpha$-maps obtained from feeding two samples of the Cityscapes validation set to the semantic fusion gate, projected and visualized as heat maps. The brighter a region is in the figure, the more relevant it is to the task. In both of the illustrated examples, it is possible to observe that pixels located near the borders of objects are considered the brightest ones. Another noticeable point is that objects that are relatively smaller in a given scene appear to receive more attention from the boundary stream.

\begin{figure}[t]
    \centering
    % \subfigure[]{\includegraphics[width=.17\textwidth]{figs/quantitative/frankfurt_000000_002196_leftImg8bit_crop.png}\label{fig:eval_quantitative_results:a}}
    % \subfigure[]{\includegraphics[width=.17\textwidth]{figs/quantitative/frankfurt_000000_002196_gtFine_color_crop.png}\label{fig:eval_quantitative_results:b}} \\
    \subfigure[]{\includegraphics[width=.17\textwidth]{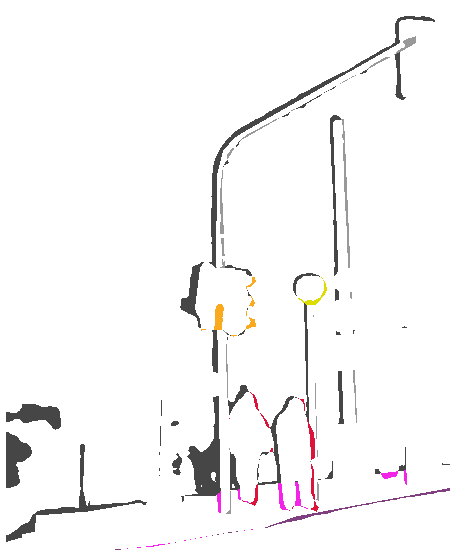}\label{fig:eval_quantitative_results:a}}
    \subfigure[]{\includegraphics[width=.17\textwidth]{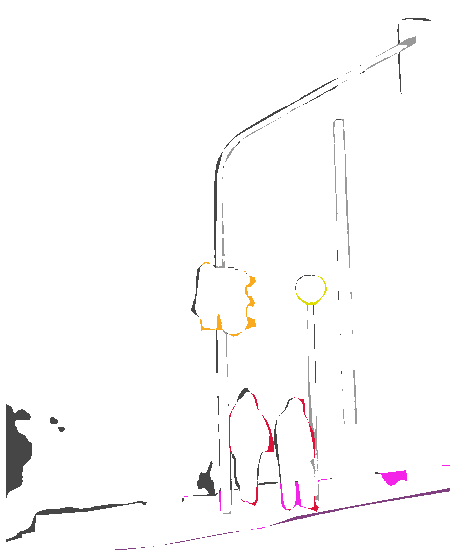}\label{fig:eval_quantitative_results:b}}

    \caption{Comparison of the prediction segmentation error on the boundaries with respect to the ground truth: (a) DeepLabV3~\citep{deeplabv3}, (b) our Y-model.}
    \label{fig:eval_quantitative_results}
\end{figure}

\paragraph{\textbf{Boundary improvement}}
\Figref{fig:eval_quantitative_results} shows the results of how our model can improve semantic segmentation on boundary regions. To see a visual summary of the error of a semantic segmentation model, we take that model prediction from a given sample and subtract it from the ground truth. We perform these steps using DeepLabV3 and our proposed model to obtain \Figref{fig:eval_quantitative_results:a} and \Figref{fig:eval_quantitative_results:b}, respectively. The results lead us to two conclusions: First, that our Y-model is capable of further refining segmentation borders when compared with other similar methods, and second that the gains reported in \Figref{fig:eval_quantitative_results} are obtained mainly through the improvement of boundary detection.

% \paragraph{Visual comparison with the ground truth} 
% We used the Cityscapes validation set to perform a simple side-by-side comparison of the results obtained by our Y-model and the respective ground truth from each sample. Like the criterion of the selection, we choose some of the best results. \Figref{fig:eval_quantitative_results_cityscapes_w} illustrates this experiment with the leftmost column being the raw sample images, the middle column being the ground truth, and the rightmost column being the results obtained from our Y-model. It is noteworthy that the results are very close to the ground truth annotation, notably presenting clear and defined boundaries for most objects in each scene.

% \begin{table}[t]
% \centering
% \caption{Comparative results on the Pascal VOC 2012 testing set.}
% % \resizebox{1\linewidth}{!}
%  {
%  \begin{tabular}{l | c}
%  \hline
%  {\bf Method } & {\bf mIoU (\%) } \\ [0.5ex] 
%   \hline
%   \hline
%  Ours (dilated-ResNet101) & 82.3  \\ [0.5ex]
%  RefineNet~\citep{yu2015multi}  &   83.4 \\ 
%  PSPNet (dilated-ResNet101)~\citep{pspnet}  & 85.4 \\
%  DeepLabV3+(Xception)~\citep{deeplabv3plus} & {\bf 85.7} \\
%  DeepLabV3 (dilated-ResNet101)~\citep{deeplabv3} -  & {\bf 85.7} \\
%  \hline
%  \end{tabular} }
%  \label{tab:comp_sota_pascal_voc}
% \end{table}

\subsection{Discussion}
\paragraph{\textbf{Architecture}} We explored the semantic boundary detection to improve the overall results of the semantic segmentation. For that, a semantic fusion gate was conceived to split semantic segmentation information between two convolutional network streams. Our Y-model structure works by capturing the different characteristics between the boundary and non-boundary pixels, and it was capable of combining both to perform the semantic segmentation task. This conclusion is backed by the results reported for the experiments carried out on different publicly available data sets that showed that our proposed model performs adequately. Tables \ref{tab:comp_sota_cityscapes}, \ref{tab:comp_sota_camvid}, and \ref{tab:comp_sota_pascal_context} clearly show that our Y-model was able to improve results when compared to other approaches in the literature. Even when the result achieved was not the best, the network was still competitive, as shown in Table \ref{tab:comp_sota_mapillary}. As shown in \Figref{fig:eval_quantitative}, the current state-of-the-art models have issues to properly classify pixels on the image object boundary, and so we attribute our best results mainly to the improved representation and integration of the boundary features. Splitting segmentation and boundary detection allowed our proposed model for a better learning, since the Y-model focuses on representing the context of pixels independently of the borders. 

\paragraph{\textbf{Semantic fusion gate}} Each stream (boundary and non-boundary) that composes our Y-model is responsible for specializing in different types of information for image segmentation. At the same time, however, these streams share enough information as to complement each other. This sharing is made possible through the semantic fusion gate, which we demonstrated to be an effective manner to fuse information from the two different sub-networks (streams). Table \ref{tab:comp_different_type_sbcnn} showed that when using attention gating the results are generally better in comparison with other models that do not apply this merger by attention. The attention gate strategy showcases the importance of fusing features from the boundary and non-boundary streams in a witting way.

\paragraph{\textbf{Integral crop}} The data sets used in the experiments were burden by imbalance. This fact makes the less prevalent classes more difficult to learn. The proposed integral crop strategy aimed at reducing this effect by assigning weight values to each pixel by considering the inverse frequency of each class over the whole data set. This is so to hopefully select regions that best represents an uniform distribution of the data during the training stage. The results in Table \ref{tab:comp_baseline_integral} showed that this strategy could improve the segmentation results. Yet the use of integral images to compute each crop weight made the algorithm possible to run in linear time -- mainly important as to avoid excessive pre-processing overhead during training.

\section{Concluding remarks}

Our proposed model has achieved results superior to the current state-of-the-art on most of the evaluated data sets. The promising results found ought to motivate other studies that explicitly explore the use of semantic boundaries for image segmentation in convolutional networks. We showed that explicitly considering semantic boundaries as part of the problem modeling may allow for a more satisfactory adjustment of the segmentation output with regards to the boundaries of objects in the scene. Our semantic fusion gate facilitated the learning of the two tasks simultaneously but separately, with each stream focusing on the appropriate information. It is important to note that the explicit representation of semantic boundaries increases the computational complexity of the model in memory requirements. So elaborating on more efficient forms of memory representation for semantic boundary can also be an important path for future research. Another possibly worthy investigation is on the use of other types of attention mechanisms, like the ones based on key, query, and value functions designed to work with attention gate, for example.

% \section*{Acknowledgements}

% The authors acknowledge the National Laboratory for Scientific Computing (LNCC/MCTI, Brazil) for providing HPC resources of the SDumont supercomputer (URL: http://sdumont.lncc.br), which have contributed to the research results reported here.

% \section*{Funding}
% Luciano Oliveira has a research scholarship supported by the Conselho Nacional de Desenvolvimento Científico e Tecnológico (CNPq) under grant number 307550/2018-4.
% \section{References}
\bibliographystyle{model2-names}
\bibliography{mybibfile}

\end{document}